\documentclass[10pt,conference,a4paper]{IEEEtran}
\usepackage{cite}
\usepackage{hyperref}
\usepackage{amssymb}

\usepackage[pdftex]{graphicx}
\DeclareGraphicsExtensions{.pdf,.jpeg,.png,.eps}
\usepackage[outdir=./]{epstopdf}
\usepackage{multirow}

\ifCLASSINFOpdf
\else
\fi
\usepackage{amsmath}
\usepackage{algorithm}
\usepackage{algorithmic}
 \usepackage[caption=false,font=footnotesize]{subfig}
\usepackage{url}

\usepackage[a4paper,top=54pt,bottom=113pt,left=37pt,right=37pt]{geometry}

\begin{document}
%
\title{Active Learning Using Uncertainty Information}




%
\author{\IEEEauthorblockN{Yazhou Yang\IEEEauthorrefmark{1} \IEEEauthorrefmark{2},
Marco Loog\IEEEauthorrefmark{1} \IEEEauthorrefmark{3}}
\IEEEauthorblockA{\IEEEauthorrefmark{1}Pattern Recognition Laboratory, Delft University of Technology, Delft, The Netherlands}
\IEEEauthorblockA{\IEEEauthorrefmark{2}College of Information System and Management, National University of Defense Technology, Changsha, China}
\IEEEauthorblockA{\IEEEauthorrefmark{3}The Image Section, University of Copenhagen, Copenhagen, Denmark}
Email: \{y.yang-4, m.loog\}@tudelft.nl}


\maketitle

\begin{abstract}
Many active learning methods belong to the retraining-based approaches, which select one unlabeled instance, add it to the training set with its possible labels, retrain the classification model, and evaluate the criteria that we base our selection on. However, since the true label of the selected instance is unknown, these methods resort to calculating the average-case or worse-case performance with respect to the unknown label. In this paper, we propose a different method to solve this problem. In particular, our method aims to make use of the uncertainty information to enhance the performance of retraining-based models. We apply our method to two state-of-the-art algorithms and carry out extensive experiments on a wide variety of real-world datasets. The results clearly demonstrate the effectiveness of the proposed method and indicate it can reduce human labeling efforts in many real-life applications.
\end{abstract}


%
\IEEEpeerreviewmaketitle

\section{Introduction}
Over the past decade, a primary foundation of much progress in machine learning is the rapid growth of the number and size of data sets available, such as ImageNet \cite{imagenet_cvpr09} containing over 14 million labeled images for object recognition. In a practical scenario, we frequently encounter the situation where few labeled instances along with abundant unlabeled samples are available. Labeling a large amount of data is, however, very difficult due to the huge amount of time required or expensive because of the need of human experts \cite{settles2010active}. Thus, it is very attractive to propose a proper labeling scheme to reduce the number of labels required in order to train a classifier.

Active learning has been put forward to overcome the above labeling problem. The main assumption behind active learning is that if an active learner can freely select any samples it wants, it can outperform random sampling with less labeling  \cite{settles2010active}. Thus, the main task of active learning is querying as little data as possible to minimize the annotation cost while maximizing the learning performance. Active learning tries to achieve this by selecting the most valuable samples. However, it is difficult to define or measure the value of one instance to the learning problem. We can view it as the amount of information carried which potentially promotes the  learning performance, once its true label is known \cite{chattopadhyay2013batch}. As a result of the fact that we do not have an exact measure of the value, there are a great number of selection criteria proposed from different perspectives on how to estimate the usefulness of each sample.  

Most commonly used criteria in active learning include query-by-committee \cite{seung1992query}, uncertainty sampling  \cite{lewis1994heterogeneous, tong2002support, campbell2000query}, expected error reduction \cite{roy2001toward, guo2007optimistic, holub2008entropy, guo2008discriminative}, expected model change \cite{settles2008multiple, freytag2014selecting, cai2014active, kading2015active}, variance reduction \cite{hoi2006batch, zhang2000value, yu2006active, schein2007active} and ``Min-max'' view active learning \cite{hoi2008semi,huang2010active}. Query-by-committee put forward multiple models as the committees and selected the samples which receive highest level of disagreement from the committees   \cite{seung1992query}. Uncertainty sampling approach preferred the instances with maximum uncertainty. Based on the measurement of uncertainty, uncertainty sampling can be roughly divided two categories: maximum entropy of the estimated label  \cite{lewis1994heterogeneous} and minimum distance from the decision boundary \cite{tong2002support, campbell2000query}. For example, Tong and Koller \cite{tong2002support} proposed to query the instance which is closed to the current learning boundary using the classifier of support vector machines. Campbell \emph{et al.}  \cite{campbell2000query} shared the same idea with Tong and Koller \cite{tong2002support}.  

Roy and McCallum \cite{roy2001toward} proposed the expected error reduction (EER), which is a popular active learning method. EER aimed to reduce the  generalization error when labeling a new instance. Since we do not have access to the test data, Roy and McCallum suggested to  compute the ``future error'' on the unlabeled pool  under the assumption that the unlabeled data set is representative of the test distribution. In other words, the unlabeled pool can be viewed as a validation set. Also, we  have no knowledge about the true labels of unlabeled samples. EER estimated the average-case criterion of potential loss instead. Expected model change followed the idea of EER, but turned to select the instance which leads to maximum change of the current model. The variance reduction methods tried to minimize the output variances \cite{settles2010active}. Schein and Ungar \cite{schein2007active} extended this approach to expected variance reduction method on logistic regression by following the idea of EER. ``Min-max'' view active learning was originally proposed by Hoi \emph{et al.} \cite{hoi2008semi}, where ``Min-max'' indicates the worst-case criterion is adopted. The key idea behind is to select the sample which minimizes the gain of objective function no matter what its assigned label is.  Huang \emph{et al.} \cite{huang2010active} extended this framework by taking into account all the unlabeled data when calculating the objective function.

Current active learning methods can be split in two classes: retraining-based and retraining-free active learning. Retraining-based active learning represents methods which measure the information of unlabeled sample by labeling it (any possible label) and adding it to the training set to retrain the classification model. Then, some appropriate criteria can be evaluated and used for the sample selection. The second class, retraining-free active learning, contains the remaining methods which not need repeatedly train the model for each unlabeled instance during one single selection. For example, uncertainty sampling and query-by-committee belong to this category. 

However, since the true label of the selected unlabeled instance is unknown, these methods resort to calculating the average-case or worse-case criteria with respect to the unknown label. In this paper, we propose a different criterion for retraining-based methods. We incorporate the uncertainty information (measured by the posterior probabilities within the min-max framework) for the selection. The proposed criterion can be seen as a trade-off of the exploration and the exploitation. The uncertainty information plays the role of the exploitation while the retraining-based models act as the exploration part. We concentrate on the pool-based active learning setting which assumes a large pool of unlabeled data along with a small set of labeled data already available \cite{settles2010active}. We consider the myopic active learning which sequentially and iteratively selects unlabeled instance. 

\subsection{Outline}
The rest of this paper is organized as follows.  Section \ref{related_work} firstly reviews the framework of retraining-based active learning. Then two state-of-the-art methods under the retraining framework are briefly described. Section \ref{proposed_framework} demonstrates the primary motivation of the proposed method and derives a general algorithm for retraining-based active learning in detail. It also illustrate how to extend the proposed criterion to current methods. Experimental design and results are reported in \ref{experiment_design} ; Section \ref{conclusions} concludes this work followed by some future issues.
\section{Retraining-based Active Learning}
\label{related_work}
In this section, we summarize a general framework of retraining-based active learning. Then we demonstrate two examples under this framework: Expected error reduction and Minimum Loss Increase.
\subsection{Retraining-based Active Learning }
\label{frame_work}

Firstly, let us introduce some preliminaries and notation. Let $\mathcal{L} ={\left\{ (x_i,y_i) \right\}}_{i=1}^{m} $ represent the training data set that consists of $m$ labeled instances and $ \mathcal{U} $ be the pool of unlabeled instances $ \left\{ x_i \right\} _ {i=m+1}^ {n} $ . Each $ x_i \in \mathbb{R}^{d} $ is a $d$ dimensional feature vector, and $ y_i \in C =\left\{ +1,-1 \right\} $ is the class label of $x_i$. In this paper, let us focus on binary classification problem firstly, and it is easy to extend this work to multi-class problem by extending $C$ to multi-labels set. We denote $P_{\mathcal{L}} (y|x)$ be the conditional probability of $y$ given $x$ according to a classifier trained on $\mathcal{L}$.

\begin{algorithm}[tbp]
	\caption{General Retraining-based Active Learning Procedure}
	\label{alg_activelearing}
	\begin{algorithmic}	[1]
		\STATE {\bfseries Input:} Labeled data $\mathcal{L}$, unlabeled data $\mathcal{U}$ 
		\REPEAT
		\STATE Train the classifier on $\mathcal{L}$ and calculate $P_{\mathcal{L}} (y_i|x_i)$ for each $x_i \in \mathcal{U}$, each $y_i \in C $;
		\FOR {each $x_i \in \mathcal{U}$}
		\FOR {each $y_i \in C $}
		\STATE Re-train the model on $ \mathcal{L} \cup \{ {x_i, y_i}\} $;
		\STATE Calculate some criterion $V(x_i,y_i)$, (\emph{e.g.}, error or variance);
		\ENDFOR 
		\ENDFOR
		\STATE Compute some kind of performance based on $P_{\mathcal{L}} (y_i|x_i)$ and $V(x_i,y_i)$;
		\STATE Query the instance $x^*$ which leads to the best performance and label it $y^*$, update $ \mathcal{L} \leftarrow \mathcal{L} \cup \{ {x^*, y^*}\},  \mathcal{U} \leftarrow \mathcal{U} \backslash \{ {x^*} \} $;	 
		\UNTIL{Stopping criterion is satisfied}
	\end{algorithmic}
\end{algorithm}

For the retraining-based active learning, its framework can be summarized in Algorithm \ref{alg_activelearing},  where $V(x_i,y_i)$ represents any selection criterion associated with $ (x_i,y_i)$. The main procedure contains the loops which checks all the points in unlabeled pool $\mathcal{U}$  over all the possible labels. For example, we firstly select one instance from the unlabeled pool and assign it any possible label. Then we update the labeled set (since we acquire a new labeled sample) and retrain the classifier we use. Based on the new trained classifier, we can measure some kind of selection criteria (\emph{e.g.}, generalization error in EER \cite{roy2001toward}). However, since the true label information of last selected sample is unknown, we need calculate some kind of performance, \emph{e.g.}, the average-case in \cite{roy2001toward, schein2007active, freytag2014selecting}, worst-case in \cite{huang2010active}, or even the best-case criteria in \cite{guo2007optimistic}. Finally, we will query the instance which leads to maximum or minimum value in terms of the criterion we are interested in.

EER is one example of retraining-based active learning, which uses the generalization error as $V(x_i,y_i)$. We get expected model change \cite{settles2008multiple, freytag2014selecting, cai2014active, kading2015active} by adopting model change as the criterion. By adopting variance and logistic regression as the classifier, we get expected variance reduction \cite{schein2007active}. Similarly, if we want to minimize the value of objective function after labeling a new instance and use the worst-case performance (corresponding to min-max framework), then we can get  \cite{hoi2008semi, huang2010active}. Clearly, the retraining-based approaches may suffer from high computational cost due to the fact that they need go over all the unlabeled data and all the possible labels. 

\subsection{Expected Error Reduction }
\label{imp_eer}
Expected error reduction has demonstrated its effectiveness on text classification domain \cite{roy2001toward}. There are also some follow-up work of EER contributed by other researchers \cite{guo2007optimistic} \cite{holub2008entropy} \cite{guo2008discriminative}. EER aims to select the sample which will reduce the future generalization error.  Since we can not see the test data, the unlabeled pool can be used as the validation set to predict the future test error. We encounter a new problem since we do not know the true labels of the pool. Roy and McCallum \cite{roy2001toward} suggested, in practice,  we can approximately estimate the error using the expected log-loss or $0/1$ loss over the pool. For example, if we adopt the log loss, EER can be written as follows:
\begin{small}
	\begin{equation*}\label{avg-eer}
	\arg \min_{x \in \mathcal{U}} \sum_{y \in C} P_{\mathcal{L}}(y|x) \left( -\sum_{x_i \in \mathcal{U}}  \sum_{y_i \in C}P_{\mathcal{L}^{+}}(y_i|x_i) \log P_{\mathcal{L}^{+}}(y_i|x_i)  \right)
	\end{equation*}
\end{small}
where $ \mathcal{L}^{+} = \mathcal{L}\cup(x,y)$ means that the selected instance $x$ is labeled $y$ and added to $\mathcal{L}$. Note that the first term $P_{\mathcal{L}}(y|x)$ contains the pre-trained label information. The second term is the sum of potential entropy over the unlabeled data set $\mathcal{U}$.

\subsection{Minimum Loss Increase}
\label{general loss}
We can find that EER attempts to reduce the future generalization error, however, it is not easy due to the missing of test data and true label information of unlabeled data. There are some researchers which try to solve this problem from a different perspective. Hoi \emph{et at.} \cite{hoi2008semi} presented a so called ``min-max'' view active learning. It prefers the instance which results in a small value of an objective function in spite of its assigned label. This is because the smaller the value of an objective function, the better the learning model, at least in high probability. Assume $G_{\mathcal{L}}$ is the value of an objective function on current labeled data  $\mathcal{L}$. When we label a new instance and update the training data $\mathcal{L^+}=\mathcal{L}\cup \{ {x_i, y_i}\}$, we get a new value of objective function $G_{\mathcal{L}^+}$. What we want is the minimum increase of objective function, \emph{i.e.,} $G_{\mathcal{L}^{+}}- G_{\mathcal{L}}$,  when adding one more labeled sample. However, because the second term $G_{\mathcal{L}}$ is independent of the next queried instance, so we can ignore it and focus on minimizing $G_{\mathcal{L}^+}$. Since we expect a minimum value of $G_{\mathcal{L}^+}$ regardless of the assigned label of $x_i$, we adopt the worst-case performance as follows, instead of the average-case version.
\begin{equation*}\label{empirical risk}
\arg \min_{x_i \in \mathcal{U}} \max_{y_i\in C} G_{\mathcal{L}^{+}}
\end{equation*}
Note that we can view $G_{\mathcal{L}^{+}}$ as one choice of $V(x_i,y_i)$ mentioned in Algorithm \ref{alg_activelearing}. 

Let us consider an unconstrained optimization problem using $L_2$-loss regularized classifier with arbitrary loss $l(w;x_i,y_i)$:
$ g(w)= \frac{1}{2\lambda} || w ||^2 + \sum_{x_i \in \mathcal{L}} l(w;x_i,y_i) $, where $w$ is the parameter of learning classifier. If we adopt the Hinge loss $l(w;x_i,y_i)=\max(0,1-y_iw^Tx_i)$, we can derive the same model with ``min-max'' view active learning described in \cite{hoi2008semi}, but without extend it to batch model setting. If we use square loss $l(w;x_i,y_i)=(y_i-w^Tx_i)^2$, we can get the same model with \cite{huang2010active}. Note that, as is stated in \cite{yang2016benchmark}, though \cite{huang2010active} includes all the unlabeled data when calculating the objective function, the unlabeled examples play no role since \cite{huang2010active} relaxes the constraint of the labels of unlabeled pool in the end. This operation can guarantee \emph{zero} contribution of unlabeled data to the objection function. Thus, \cite{huang2010active} is also one special case using the square loss. Moreover, we can conclude that the main idea of min-max view active learning is to minimize the increase of the value of an objective function. 

In our paper, we consider the logistic loss $l(w;x_i,y_i)=\log(1+\exp^{-y_iw^Tx_i})$, which results in:
\begin{small} 
	\begin{equation}\label{LRmodel}
	\arg \min_{x \in \mathcal{U}} \max_{y \in C} \frac{1}{2\lambda} || \hat{w} ||^2 + \sum_{x_i \in \mathcal{L}^{+}}  -\log P_{\mathcal{L}^{+}}(y_i|x_i)
	\end{equation}
\end{small}
where $\hat{w}$ is estimated parameter of $L_2$-regularized logistic regression model. Logistic regression is chosen as the base classifier since it is generally widely used in many fields and can output the conditional probability straightly, which can be used in active learning \cite{yang2016benchmark}. We call this method Minimum Loss Increase (MLI) in this paper.  EER tries to minimize the error on unlabeled data while MLI aims to minimize the loss on data already labeled. 

\section{A New Retraining-based Active Learner}
\label{proposed_framework}

In this section, we motivate our proposed
method and, subsequently, describe a general adaptation for retaining-based active learning models.

\subsection{Motivation}
\label{motivation}
Obviously, not knowing the true labels of the unlabeled data complicates calculating the final score of each instance in step 10 in Algorithm \ref{alg_activelearing}. 
One simple possibility is computing the average-case \cite{roy2001toward} or worst-case performance \cite{huang2010active}, or even the best-case criterion \cite{guo2007optimistic}. These choices, however, may fail to take into account some potentially valuable information: Firstly, although the average-case criterion makes use of the label distribution information $P_{\mathcal{L}} (y_i|x_i)$ already known, the expectation calculation can hide or underestimate some outstanding samples due to the re-weighting by $P_{\mathcal{L}} (y_i|x_i)$. For example, the true label of instance $x_i$ is $+1$ but the estimated $P_{\mathcal{L}} (+1 |x_i) = 0.1$, and the $V(x_i,+1)$ has a maximum value compared with other instances. Then the average-case criterion of $x_i$, namely $\sum_{y_i} P_{\mathcal{L}} (y_i|x_i)V(x_i,y_i)$, is highly likely to be surpassed by other instances. Secondly, as to the worst-case criterion, it suffers from not taking advantage of label distribution information at all. Worst-case analysis is a safe analysis since it is never underestimated. However, making no use of the available label information $P_{\mathcal{L}} (y_i|x_i)$ can lose sight of some valuable information. 

Thus, to overcome the shortcomings mentioned, a new criterion for retraining-based active learning is proposed. The main motivation is that we want to incorporate the uncertainty information (\emph{e.g.}, known label distribution information) within min-max framework for retraining-based models. The proposed criterion is therefore as follows:
\begin{equation}\label{new_criterion}
\min_{x_i \in \mathcal{U}} \max_{y_i \in C} P_{\mathcal{L}}(y_i|x_i) V(x_i,y_i)
\end{equation}
where $ P_{\mathcal{L}}(y_i|x_i)$ contains the pre-trained label information and $V(x_i,y_i)$ represents any criteria we are interested. Note that for some classifiers like logistic regression, we can use the estimated posterior probability as  $ P_{\mathcal{L}}(y_i|x_i)$. For classifiers which do not produce a probabilistic output, \emph{e.g.}, SVMs, we can transform their output to some probability using Platt's \cite{platt1999probabilistic} or Duin \& Tax's method \cite{duin1998classifier}. And for $V(x_i,y_i)$, various choices are possible, such as the test error on the unlabeled pool in EER, the output variance as in \cite{schein2007active}, or the value of an objective function \cite{huang2010active}.

The proposed method can be interpreted as follows: it utilizes the pre-trained label information, although this kind of information might be inaccurate due to limited labeled data we have, it still shows some underlying or potential useful clues which may promote active learning. Firstly, it improves upon the average-case criterion since it does not compute the expected value. The calculation of expectation tends to ruin the discriminative information contained in the data due to its averaging manner.  Secondly, it outperforms the worst-case criterion because it takes advantage of the knowledge of the potential label distribution while worst-case analysis does not use this at all. Thus, it avoids the disadvantages of average-case and worst-case criteria. It can be seen as a trade-off between the average-case and the worst-case criteria. Lastly, it can be considered as incorporating uncertainty sampling (encoded by the posterior probabilities) for retraining-based model. If all $V(x_i,y_i)$  become one constant term like 1 or $ P_{\mathcal{L}}(y_i|x_i)$ itself, then the proposed method will turn into exactly the uncertainty sampling. More specifically,  $\min_{x_i \in \mathcal{U}} \max_{y_i \in C} P_{\mathcal{L}}(y_i|x_i) $ or $\min_{x_i \in \mathcal{U}} \max_{y_i \in C} [P_{\mathcal{L}}(y_i|x_i)]^2 $ will act as totally same as uncertainty sampling since they will select the instance whose posterior probability comes closest to 0.5 on the binary problem. This shows that our proposed method actually fuses uncertainty sampling with retraining-based models. 
\subsection{Two Examples of the Proposed Method}
\label{applying}

\begin{figure}[tp]
	\setlength{\abovecaptionskip}{-0.3cm}
	\setlength{\belowcaptionskip}{-1cm} 
	\centering
	\includegraphics[width=\columnwidth ]{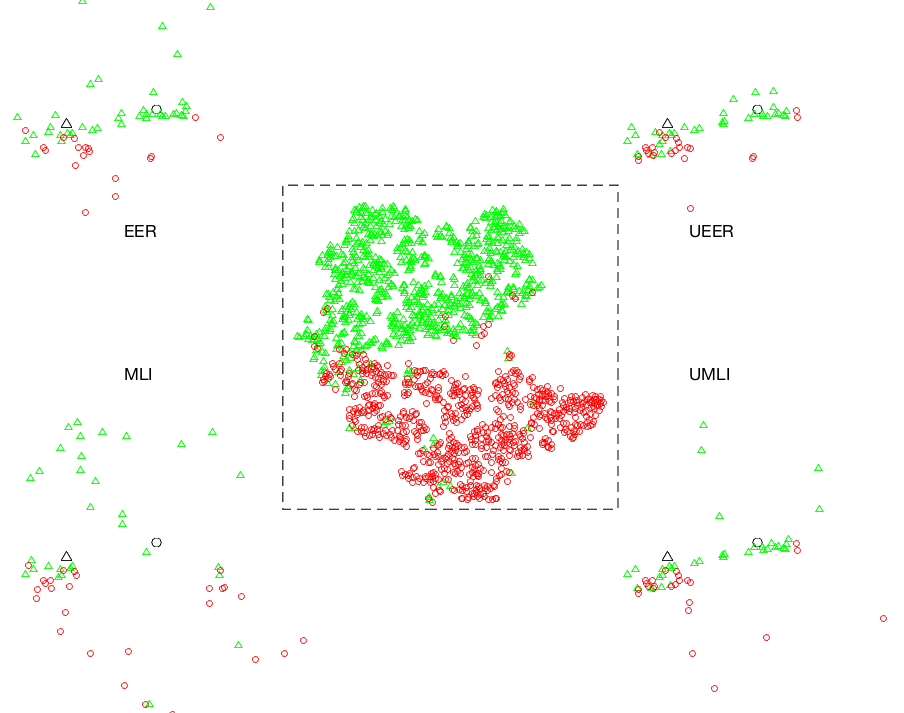}
	\caption{Illustration of the inherent characteristics of the proposed method. The middle is the distribution of a synthetic binary data set. Four corners represent the performance of four active learning methods, EER, UEER, MLI and UMLI, respectively. One black triangle and circle represent the initial labeled set.} 
	\label{figure1}
\end{figure}

To provide valuable insights on the underlying characteristic of the proposed method, we apply it to two state-of-the-art retraining-based models EER and MLI. We also demonstrate its advantage on a synthetic data set in Figure \ref{figure1}.

Since our method tries to make use of the uncertainty information, the following adapted methods are termed uncertainty retraining-based active learners. It is easy to extend EER to uncertainty-based error reduction by adopting our method in Equation \ref{new_criterion} as follows:
\begin{small}
	\begin{equation*}\label{u-eer}
	\arg \min_{x \in \mathcal{U}} \max_{y \in C} P_{\mathcal{L}}(y|x) \left( -\sum_{x_i \in \mathcal{U}} \sum_{y_i \in C}P_{\mathcal{L}^{+}}(y_i|x_i) \log P_{\mathcal{L}^{+}}(y_i|x_i)  \right)
	\end{equation*}
\end{small}
This method is called UEER for short. We can also apply our proposed criterion on MLI. The new approach is called UMLI in this paper. Note that the regularization parameter  $\frac{1}{2\lambda}$ in Equation \ref{LRmodel} is usually quite small, so we ignore it in our adapted criterion:
\begin{equation*}\label{ULRmodel}
\arg \min_{x \in \mathcal{U}} \max_{y \in C} P_{\mathcal{L}}(y|x) \! \sum_{x_i \in \mathcal{L}^{+}} \! -\log P_{\mathcal{L}^{+}}(y_i|x_i)
\end{equation*}
As is shown in Figure \ref{figure1} , we construct a synthetic binary data set and two colours represent
different classes. We demonstrate the performance of four retraining-based active learners EER, UEER, MLI and UMLI on four corners, respectively. One black triangle and circle in each corner represent two initial labeled points. When we compare UEER with EER, it is obvious that UEER selects a number of instances near the decision boundary while EER explores points in a wider range. This is because our method helps UEER make use of the uncertainty information and uncertainty information makes UEER focus on the region which is least certain about. Similar results can also be found between UMLI and MLI. MLI explores over the data space and queries the points around the	border while UMLI balances the exploration and the exploitation. UMLI concentrates on the central part (exploitation) and also searches around the edge. Therefore, we can see that our method enhances retraining-based model by balancing the exploration and the exploitation.

\section{Experiments}\label{experiment_design}
In this section, we investigate the performance of our proposed methods to examine the effectiveness and robustness of our new criterion. The following experiments are limited to binary classification problems. Firstly, we show the experimental setting, then present the extensive experiment results, followed by further discussion and analysis.
\subsection{Experimental setting}
We compare the our proposed methods UEER and UMLI against their original version EER and MLI, respectively. Random sampling is also included in this comparison. In all the experiments, we use $L_2$-regularized logistic regression included in LIBLINEAR package \cite{fan2008liblinear} as default classifier with the same regularization parameter, $\lambda = 100$, for all methods.

The classification accuracy is used as the comparison criterion in our experiment. However, since active learning is a iteratively labeling procedure, we care about the performance during the whole learning process. Thus, it is not reasonable to merely compare the accuracy at some single points. Instead, we generate the learning curve of classification accuracy versus the number of labeled instances. Then, we calculate the area under the learning curve (ALC) as a measure of evaluation.

We test on totally 49 real-world data sets from various real-life applications, including many UCI data sets \cite{Lichman2013}, MNIST handwritten digit dataset \cite{lecun1998gradient} and 20 Newsgroups dataset \cite{Lang95}. There are 39 datasets from UCI benchmark datasets, such as breast, vehicle, heart and so on. These datasets are pre-processed according to \cite{fernandez2014we}. For wine data set, we conduct class 2 against class 1 and 3 as binary problem. For glass data set, we also split it into two groups (class 1-3 vs. class 5-7) to build binary case. 
We randomly sub-sample 1000 instances from mushroom for computing efficiency. We select six pairs of letters from Letter Recognition Data Set \cite{Lichman2013}, \emph{i.e.,} D vs. P, E vs. F, I vs.J , M vs.N, V vs. Y and U vs. V since these pairs look similar to each other and distinguishing them is a little challenging. 3 vs. 5, 5 vs. 8 and 7 vs. 9 are three difficult pairs taken from MNIST data set \footnote{\href{http://yann.lecun.com/exdb/mnist/}{http://yann.lecun.com/exdb/mnist/}} and used as the binary classification data set. We randomly sub-sample 1500 instances from the three data sets for computing efficiency. We also test the performance on 20 Newsgroups dataset which is a common benchmark used for text classification \footnote{\href{http://qwone.com/~jason/20Newsgroups/}{http://qwone.com/~jason/20Newsgroups/}}. Following the work of \cite{zhu2003combining}, we also evaluate three binary tasks from 20 Newsgroups dataset: baseball vs. hockey, pc vs. mac, and religion.misc vs. alt.atheism. And the three pairs represent easy, moderate and difficult classification problems, respectively. We apply PCA to reduce the dimensionality of the above three datasets to 500 for computation efficiency. We also use the pre-processed data autos, motorcycles, baseball, hockey used in \cite{yu2006active}.

To objectively evaluate the performance, each data set is randomly divided into training and test data set of equal size. At the very beginning of active learning, we assume that only two instances randomly picked up from the training data are labeled, and one of them is from the positive class and the other is from the negative class. We run each active learning algorithm 20 times on each real-world dataset. The average performance of each active learning method is reported in the following section. 

\begin{table}[tbp]
	\setlength{\abovecaptionskip}{-0.2cm}
	\renewcommand{\arraystretch}{1.3}
	\caption{Data sets information: It shows the number of instances (\# INS) and the feature dimensionality (\# FEA)}
	\label{datasetlist}
	\begin{center}
		\resizebox{\columnwidth}{!}{ 
			\begin{tabular}{l|l|l}
				\hline
				Data set (\# Ins, \# Fea) &Data set (\# Ins, \# Fea) & Data set (\# Ins, \# Fea) \\
				\hline
				ac-inflam (120, 6)& acute (120, 6)& australian (690, 14) \\
				blood (748, 4)& breast (683, 10)& credit (690, 15) \\
				cylinder (512, 35)& diabetes (768, 8)& fertility (100, 9) \\
				german (1000, 24)& glass (214, 9)& haberman (306, 3) \\
				heart (270, 13)& hepatitis (255, 19)& hill (606, 100) \\
				ionosphere (351, 34)& liver (345, 6)& mushrooms (1000, 112) \\
				mammographic (961, 5)&  musk1 (476, 166)& ooctris2f (912, 25) \\
				ozone (1000, 72)& parkinsons (195, 22)& pima (768, 8) \\
				planning (182, 12)& sonar (208, 60)& splice (1000, 60) \\
				tictactoe (958, 9)& vc2 (310, 6)& vehicle (435, 18) \\
				wine (178, 13)& wisc (699, 9)& wdbc (569, 31) \\
				d vs p (1608, 16)& e vs f (1543, 16)&  i vs j (1502, 16) \\
				m vs n (1575, 16)&  v vs y (1577, 16)& u vs v (1550, 16) \\
				3 vs 5 (1500, 784)& 5 vs 8 (1500, 784)& 7 vs 9 (1500, 784) \\
				base-hockey (1993, 500)& pc-mac (1945, 500)& misc-atheism (1427, 500) \\
				autos (3970, 8014)& motorcycles (3970, 8014)& baseball (3970, 8014) \\
				hockey (3970, 8014) & &\\ 
				\hline
			\end{tabular}
		}
	\end{center}
\end{table}

\subsection{Results}

\begin{table}[htb]
	\setlength{\abovecaptionskip}{-0.1cm}
	\caption{Performance Comparison on the areas under the learning curve (ALC)}
	\label{table-accuracy}
	\begin{center}
		\resizebox{\columnwidth}{!}{ 						
			\begin{tabular}{l|c | cc|cc }
				\hline
				Dataset & Random  & EER & UEER &  MLI &UMLI  \\
				\hline
				hill & 0.581 & \textbf{0.616} & 0.599 & \textbf{0.626} & 0.612 \\ 
				planning & 0.586 & \textbf{0.58} & 0.578 & \textbf{0.614} & 0.586 \\ 
				cylinder & 0.586 & \textbf{0.61} & 0.597 & 0.608 & \textbf{0.617} \\ 
				liver & 0.627 & \textbf{0.635} & 0.626 & \textbf{0.615} & 0.607 \\ 
				splice & 0.659 & 0.679 & \textbf{0.682} & 0.65 & \textbf{0.666} \\ 
				german & 0.664 & 0.673 & \textbf{0.679} & 0.691 & \textbf{0.703} \\ 
				ooctris2f & 0.679 & \textbf{0.678} & 0.673 & \textbf{0.686} & 0.663 \\ 
				musk1 & 0.682 & 0.699 & \textbf{0.71} & \textbf{0.702} & 0.688 \\ 
				fertility & 0.693 & 0.706 & \textbf{0.712} & \textbf{0.727} & 0.711 \\ 
				haberman & 0.711 & 0.712 & \textbf{0.715} & 0.694 & \textbf{0.7} \\ 
				sonar & 0.713 & \textbf{0.715} & 0.707 & 0.708 & \textbf{0.712} \\ 
				pima & 0.716 & 0.706 & \textbf{0.714} & 0.711 & \textbf{0.722} \\ 
				pcmac & 0.717 & \textbf{0.715} & \textbf{0.716} & 0.747 & \textbf{0.751} \\ 
				diabetes & 0.719 & \textbf{0.723} & \textbf{0.723} & \textbf{0.726} & \textbf{0.728} \\ 
				religionatheism & 0.72 & 0.708 & \textbf{0.718} & 0.691 & \textbf{0.739} \\ 
				hepatitis & 0.731 & \textbf{0.753} & \textbf{0.75} & 0.73 & \textbf{0.738} \\ 
				blood & 0.743 & \textbf{0.74} & 0.718 & 0.73 & \textbf{0.732} \\ 
				baseball & 0.753 & 0.765 & \textbf{0.872} & 0.832 & \textbf{0.847} \\ 
				motorcycles & 0.763 & 0.78 & \textbf{0.883} & 0.854 & \textbf{0.859} \\ 
				autos & 0.768 & 0.768 & \textbf{0.872} & \textbf{0.838} & \textbf{0.835} \\ 
				heart & 0.774 & 0.791 & \textbf{0.795} & \textbf{0.797} & \textbf{0.799} \\ 
				hockey & 0.775 & 0.787 & \textbf{0.901} & 0.875 & \textbf{0.882} \\ 
				ionosphere & 0.779 & \textbf{0.818} & 0.806 & 0.674 & \textbf{0.766} \\ 
				credit & 0.779 & 0.793 & \textbf{0.814} & 0.797 & \textbf{0.809} \\ 
				mammographic & 0.78 & 0.774 & \textbf{0.795} & 0.766 & \textbf{0.779} \\ 
				basehockey & 0.793 & 0.785 & \textbf{0.801} & 0.817 & \textbf{0.847} \\ 
				vc2 & 0.807 & \textbf{0.815} & 0.812 & \textbf{0.825} & 0.82 \\ 
				parkinsons & 0.811 & \textbf{0.824} & \textbf{0.821} & \textbf{0.83} & 0.826 \\ 
				australian & 0.823 & 0.832 & \textbf{0.84} & \textbf{0.842} & 0.83 \\ 
				letterIJ & 0.853 & \textbf{0.879} & 0.853 & 0.865 & \textbf{0.874} \\ 
				letterVY & 0.855 & 0.878 & \textbf{0.884} & 0.861 & \textbf{0.867} \\ 
				3vs5 & 0.856 & \textbf{0.903} & 0.897 & 0.859 & \textbf{0.872} \\ 
				vehicle & 0.859 & 0.878 & \textbf{0.888} & 0.883 & \textbf{0.89} \\ 
				5vs8 & 0.864 & \textbf{0.907} & 0.901 & 0.85 & \textbf{0.87} \\ 
				7vs9 & 0.876 & 0.914 & \textbf{0.921} & 0.841 & \textbf{0.874} \\ 
				ozone & 0.882 & 0.86 & \textbf{0.899} & \textbf{0.892} & 0.882 \\ 
				tictactoe & 0.894 & \textbf{0.912} & 0.899 & 0.853 & \textbf{0.88} \\ 
				glass & 0.904 & \textbf{0.914} & \textbf{0.914} & \textbf{0.917} & 0.912 \\ 
				wine & 0.906 & 0.936 & \textbf{0.943} & \textbf{0.94} & \textbf{0.939} \\ 
				letterMN & 0.916 & \textbf{0.944} & \textbf{0.941} & 0.927 & \textbf{0.932} \\ 
				mushrooms & 0.931 & 0.969 & \textbf{0.974} & \textbf{0.971} & \textbf{0.972} \\ 
				letterEF & 0.933 & 0.954 & \textbf{0.961} & 0.956 & \textbf{0.957} \\ 
				wdbc & 0.938 & 0.953 & \textbf{0.956} & \textbf{0.958} & \textbf{0.957} \\ 
				letterDP & 0.939 & 0.963 & \textbf{0.969} & \textbf{0.967} & \textbf{0.966} \\ 
				letterUV & 0.945 & 0.972 & \textbf{0.979} & \textbf{0.974} & \textbf{0.974} \\ 
				wisc & 0.949 & 0.951 & \textbf{0.954} & 0.956 & \textbf{0.956} \\ 
				breast & 0.95 & 0.956 & \textbf{0.959} & \textbf{0.962} & \textbf{0.962} \\ 
				ac-inflam & 0.955 & \textbf{0.981} & \textbf{0.984} & \textbf{0.98} & \textbf{0.983} \\ 
				acute & 0.978 & 0.971 & \textbf{0.984} & \textbf{0.992} & \textbf{0.992} \\  
				\hline
				Mean & 0.798 & 0.812 & \textbf{0.822} & 0.812 & \textbf{0.818} \\
				Average Rank & 4.143 & 3.102 &\textbf{ 2.388} & 2.857 & \textbf{2.510} \\ 
				\hline \hline
				Win/tie/loss counts & - & \multicolumn{2}{c|}{29/7/13}  & \multicolumn{2}{c}{27/11/11} \\
				\hline
			\end{tabular}
		}
	\end{center}
\end{table}

Table~\ref{table-accuracy}  shows the experimental results on 49 data sets. The datasets in Table \ref{table-accuracy} are sorted with respect to the performance of random sampling. We can find that the comparisons contain the datasets which vary from very difficult problems (\emph{e.g.}, hill) to easy tasks (\emph{e.g.}, acute). To clearly demonstrate the advantage of the proposed method, we do pairwise comparison between the original algorithm and its counterpart, \emph{e.g.}, EER vs. UEER and MLI vs. UMLI, respectively. On each data set, a paired t-tests at $95\%$ significance level is used to determine which method has the best performance or provides comparable outcome. These methods are highlighted in bold face. Over all the experiments, average performances are reported in Table \ref{table-accuracy}. ``Average Rank'' shows the average rank of all the methods with regard to their performances on all the experiments. The lower the value of average rank, the better the method. The ``win/tie/loss counts'' represents times of our proposed methods versus its counterparts over all the 49 datasets.

As is shown in Table~\ref{table-accuracy}, our proposed methods UEER and UMLI evidently outperform their counterparts EER and MLI, respectively. UEER surpasses EER in terms of average accuracy, and improves its performance from 0.812 to 0.822. UEER also outperforms EER in terms of ``average rank'', which demonstrates the effectiveness of our method. Similar results can be found between UMLI and MLI. UMLI is superior to MLI on the overall performance. Moreover, it is interesting to observe that UEER attains the best overall performance among all the active learning methods.  Over all the experimental data sets, the ``win/tie/loss'' counts of UEER versus EER is $29/7/13 $, meaning that UEER is the preferred active learner in over half the cases. With regard to UMLI and MLI, the ``win/tie/loss'' count is $ 27/11/11 $, which also shows the clear benefit of our scheme nonetheless. We also notice that even random sampling can surpass all the other methods, \emph{e.g.}, on the blood data set, indicating that, generally, one might not want to use active learners in a blind way.

To investigate the robustness of our method, we also apply the worst-case criterion on EER and the average-case criterion on MLI, respectively. Due to the lack of space, we omit the results on each data set and only report the average performances. The average performance (ALC) of the worst-case on EER is 0.771 while that of the average-case on MLI is 0.710. To our surprise, they definitely show poorer performances in comparison with our method and even perform worse than random sampling. The possible reason may be that: EER computes the error on the unlabeled data and none of the true label are known, the average-case criterion is a safe choice for EER. Since MLI estimates the loss on the enlarged labeled set $\mathcal{L}\cup \{ {x_i, y_i}\}$ and only the true label of $x_i$ is unknown, the worst-case criterion is more appropriate for MLI than the average-case criterion. However, since the proposed method is a trade-off of the two criteria, it can adapt to both settings and show a robust performance for different retraining-based models.

%

\section{Conclusions}
\label{conclusions}
In this paper, we propose a new general method for retraining-based active learning. The proposed method can balance a trade-off of the average-case and worst-case criteria by incorporating uncertainty information (carried by the pre-trained posterior probabilities) within min-max framework. It drives current retraining-based models to pay more attention to the exploitation. We employ the new idea on two state-of-the-art methods to investigate its effectiveness. The synthetic data demonstrates that our method prefers to select the instances which are near the decision boundary in comparison with the original retraining-based approaches. Moreover, extensive experiments on 49 real-world datasets also prove that the proposed method is a promising approach for promoting retraining-based active learners.

\bibliographystyle{IEEEtran}
\bibliography{IEEEabrv,mybibfile}

\end{document}